\documentstyle[12pt,mkstyle,twoside]{article}
\newcommand{\ML}[1]{}     
\newcommand{\NS}[1]{#1}   
\NS{\title{p-d-Separation - A Concept for Expressing
Dependence/Independence Relations in Causal Networks}
\shorttitle{p-d-Separation in Causal Networks}
}

\newcommand{\Bem}[1]{}

\newcommand{\AbbEins}
{
\begin{figure}
\begin{center}
\begin{picture}(8,8.3)
\put(4.7,2){\circle*{0.2}} 
\put(4.4,1.3){$X_7$}
\put(4.9,2){\vector(1,0){1.3}}
\put(4.7,2){\vector(0, 1){1.3}}    
\put(4.9,2.2){\vector(1, 2){1.7}}    
\put(6.6,2){\circle*{0.2}}
\put(6.6,1.3){$X_8$}
\put(3,5.6){\circle*{0.2}}
\put(2.9,6.1){$X_4$} 
\put(2.8,5.6){\vector(-1,0){1.3}}  
\put(3.2,5.6){\vector(1,0){1.3}}
\put(4.7,7.4){\circle*{0.2}} 
\put(5,8.1){$X_{10}$} 
\put(4.5,7.2){\vector(-1,-1){1.3}}    
\put(4.7,7.2){\vector(0,-1){1.5}}
\put(4.7,5.6){\circle*{0.2}} 
\put(5,6.1){$X_9$}
\put(6.6,5.6){\circle*{0.2}}
\put(6.6,6.1){$X_5$}
\put(6.4,5.4){\vector(-1, -1){1.3}}    
\put(6.6,5.4){\vector(0,-1){3.1}}
\put(6.4,5.6){\vector(-1,0){1.3}}
\put(6.4,5.8){\vector(-1,1){1.3}}    
\put(1.2,5.6){\circle*{0.2}}
\put(1.2,6.1){$X_3$}
\put(1.2,3.8){\circle*{0.2}}
\put(1.2,4.3){$X_1$}
\put(1.4,3.8){\vector(1,0){1.3}}
\put(3,3.8){\circle*{0.2}} 
\put(2.4,4.3){$X_2$}
\put(3.2,3.8){\vector(1,0){1.3}}
\put(4.7,3.8){\circle*{0.2}} 
\put(4.4,4.3){$X_6$}
\put(4.9,3.6){\vector(1,-1){1.3}}    
\end{picture}
\end{center}
\caption{An example of a Directed Acyclic Graph (dag)}  \label{abbeins}
\end{figure}
} 
\newcommand{\AbbZwei}
{
\begin{figure}
\begin{center}
\begin{picture}(8,8.3)
\put(4.7,2){\circle*{0.2}} 
\put(4.4,1.3){$X_7$}
\put(4.9,2){\line(1,0){1.3}}
\put(4.7,2){\line(0, 1){1.3}}    
\put(4.9,2.2){\line(1, 2){1.7}}    
\put(6.6,2){\circle*{0.2}}
\put(6.6,1.3){$X_8$}
\put(3,5.6){\circle*{0.2}}
\put(2.9,6.1){$X_4$} 
\put(2.8,5.6){\line(-1,0){1.3}}  
\put(3.2,5.6){\line(1,0){1.3}}
\put(4.7,7.4){\circle*{0.2}} 
\put(5,8.1){$X_{10}$} 
\put(4.5,7.2){\line(-1,-1){1.3}}    
\put(4.7,7.2){\line(0,-1){1.3}}
\put(4.7,5.6){\circle*{0.2}} 
\put(5,6.1){$X_9$}
\put(6.6,5.6){\circle*{0.2}}
\put(6.6,6.1){$X_5$}
\put(6.4,5.4){\line(-1, -1){1.3}}    
\put(6.6,5.4){\line(0,-1){3.1}}
\put(6.4,5.6){\line(-1,0){1.3}}
\put(6.4,5.8){\line(-1,1){1.3}}    
\put(1.2,5.6){\circle*{0.2}}
\put(1.2,6.1){$X_3$}
\put(1.2,3.8){\circle*{0.2}}
\put(1.2,4.3){$X_1$}
\put(1.4,3.8){\line(1,0){1.3}}
\put(3,3.8){\circle*{0.2}} 
\put(2.4,4.3){$X_2$}
\put(3.2,3.8){\line(1,0){1.3}}
\put(4.7,3.8){\circle*{0.2}} 
\put(4.4,4.3){$X_6$}
\put(4.9,3.6){\line(1,-1){1.3}}    
\end{picture}
\end{center}
\caption{An undirected graph obtained by application of Principle I}
\label{abbzwei}
\end{figure}
} 
\newcommand{\AbbDrei}
{
\begin{figure}
\begin{center}
\begin{picture}(8,8.3)
\put(4.7,2){\circle*{0.2}} 
\put(4.4,1.3){$X_7$}
\put(4.9,2){\line(1,0){1.3}}     
\put(4.7,2){\vector(0, 1){1.3}}    
\put(4.9,2.2){\line(1, 2){1.7}}    
\put(6.6,2){\circle*{0.2}}
\put(6.6,1.3){$X_8$}
\put(3,5.6){\circle*{0.2}}
\put(2.8,5.6){\line(-1,0){1.3}}  
\put(2.9,6.1){$X_4$} 
\put(3.2,5.6){\vector(1,0){1.3}}   
\put(4.7,7.4){\circle*{0.2}} 
\put(5,8.1){$X_{10}$} 
\put(4.5,7.2){\line(-1,-1){1.3}}    
\put(4.7,7.2){\line(0,-1){1.3}}   
\put(4.7,5.6){\circle*{0.2}} 
\put(5,6.1){$X_9$}
\put(6.6,5.6){\circle*{0.2}}
\put(6.6,6.1){$X_5$}
\put(6.4,5.4){\vector(-1, -1){1.3}}    
\put(6.6,3.4){\line(0,-1){1.2}}    
\put(6.6,5.4){\line(0,-1){3.1}}   
\put(6.4,5.6){\vector(-1,0){1.3}} 
\put(6.4,5.8){\line(-1,1){1.3}}    
\put(1.2,5.6){\circle*{0.2}}
\put(1.2,6.1){$X_3$}
\put(1.2,3.8){\circle*{0.2}}
\put(1.2,4.3){$X_1$}
\put(1.4,3.8){\line(1,0){1.3}}
\put(3,3.8){\circle*{0.2}} 
\put(2.4,4.3){$X_2$}
\put(3.2,3.8){\vector(1,0){1.3}} 
\put(4.7,3.8){\circle*{0.2}} 
\put(4.4,4.3){$X_6$}
\put(4.9,3.6){\line(1,-1){1.3}}    
\end{picture}
\end{center}
\caption{A partially oriented graph due to Principle II (nodes $X_4,X_8,X_9$)}
\label{abbdrei}
\end{figure}
} 
\newcommand{\AbbVier}
{
\begin{figure}
\begin{center}
\begin{picture}(8,8.3)
\put(4.7,2){\circle*{0.2}} 
\put(4.4,1.3){$X_7$}
\put(4.9,2){\line(1,0){1.3}}     
\put(4.7,2){\vector(0, 1){1.3}}    
\put(4.9,2.2){\line(1, 2){1.7}}    
\put(6.6,2){\circle*{0.2}}
\put(6.6,1.3){$X_8$}
\put(3,5.6){\circle*{0.2}}
\put(2.8,5.6){\line(-1,0){1.3}}  
\put(2.9,6.1){$X_4$} 
\put(3.2,5.6){\vector(1,0){1.3}}   
\put(4.7,7.4){\circle*{0.2}} 
\put(5,8.1){$X_{10}$} 
\put(4.5,7.2){\line(-1,-1){1.3}}    
\put(4.7,7.2){\line(0,-1){1.3}}   
\put(4.7,5.6){\circle*{0.2}} 
\put(5,6.1){$X_9$}
\put(6.6,5.6){\circle*{0.2}}
\put(6.6,6.1){$X_5$}
\put(6.4,5.4){\vector(-1, -1){1.3}}    
\put(6.6,3.4){\line(0,-1){1.2}}    
\put(6.6,5.4){\line(0,-1){3.1}}   
\put(6.4,5.6){\vector(-1,0){3.1}} 
\put(6.4,5.8){\line(-1,1){1.3}}    
\put(1.2,5.6){\circle*{0.2}}
\put(1.2,6.1){$X_3$}
\put(1.2,3.8){\circle*{0.2}}
\put(1.2,4.3){$X_1$}
\put(1.4,3.8){\line(1,0){1.3}}
\put(3,3.8){\circle*{0.2}} 
\put(2.4,4.3){$X_2$}
\put(3.2,3.8){\vector(1,0){1.3}} 
\put(4.7,3.8){\circle*{0.2}} 
\put(4.4,4.3){$X_6$}
\put(4.9,3.6){\vector(1,-1){1.3}}    
\end{picture}
\end{center}
\caption{A partially oriented graph due to Principle II$^c$ 
(arrow  $(X_6,X_8)$)}
\label{abbvier}
\end{figure}
} 
\newcommand{\AbbFuenf}
{
\begin{figure}
\begin{center}
\begin{picture}(8,8.3)
\put(4.7,2){\circle*{0.2}} 
\put(4.4,1.3){$X_7$}
\put(4.9,2){\vector(1,0){1.3}}     
\put(4.7,2){\vector(0, 1){1.3}}    
\put(4.9,2.2){\line(1, 2){1.7}}    
\put(6.6,2){\circle*{0.2}}
\put(6.6,1.3){$X_8$}
\put(3,5.6){\circle*{0.2}}
\put(2.8,5.6){\line(-1,0){1.3}}  
\put(2.9,6.1){$X_4$} 
\put(3.2,5.6){\vector(1,0){1.3}}   
\put(4.7,7.4){\circle*{0.2}} 
\put(5,8.1){$X_{10}$} 
\put(4.5,7.2){\line(-1,-1){1.3}}    
\put(4.7,7.2){\line(0,-1){1.3}}   
\put(4.7,5.6){\circle*{0.2}} 
\put(5,6.1){$X_9$}
\put(6.6,5.6){\circle*{0.2}}
\put(6.6,6.1){$X_5$}
\put(6.4,5.4){\vector(-1, -1){1.3}}    
\put(6.6,5.4){\vector(0,-1){3.1}}  
\put(6.4,5.6){\vector(-1,0){1.3}} 
\put(6.4,5.8){\line(-1,1){1.3}}    
\put(1.2,5.6){\circle*{0.2}}
\put(1.2,6.1){$X_3$}
\put(1.2,3.8){\circle*{0.2}}
\put(1.2,4.3){$X_1$}
\put(1.4,3.8){\line(1,0){1.3}}
\put(3,3.8){\circle*{0.2}} 
\put(2.4,4.3){$X_2$}
\put(3.2,3.8){\vector(1,0){1.3}} 
\put(4.7,3.8){\circle*{0.2}} 
\put(4.4,4.3){$X_6$}
\put(4.9,3.6){\vector(1,-1){1.3}}    
\end{picture}
\end{center}
\caption{A partially oriented graph due to Principle IV 
(arrows  $(X_7,X_8)$, $(X_5,X_8)$)}
\label{abbfuenf}
\end{figure}
} 
\newcommand{\AbbSechs}
{
\begin{figure}
\begin{center}
\begin{picture}(8,8.3)
\put(4.7,2){\circle*{0.2}} 
\put(4.4,1.3){$X_7$}
\put(4.9,2){\vector(1,0){1.3}}     
\put(4.7,2){\vector(0, 1){1.3}}    
\put(4.9,2.2){\line(1, 2){1.7}}    
\put(6.6,2){\circle*{0.2}}
\put(6.6,1.3){$X_8$}
\put(3,5.6){\circle*{0.2}}
\put(2.8,5.6){\line(-1,0){1.3}}  
\put(2.9,6.1){$X_4$} 
\put(3.2,5.6){\vector(1,0){1.3}}   
\put(4.7,7.4){\circle*{0.2}} 
\put(5,8.1){$X_{10}$} 
\put(4.5,7.2){\line(-1,-1){1.3}}    
\put(4.7,7.2){\vector(0,-1){0.9}}  
\put(4.7,6.2){\line(0,-1){0.4}}    
\put(4.7,5.6){\circle*{0.2}} 
\put(5,6.1){$X_9$}
\put(6.6,5.6){\circle*{0.2}}
\put(6.6,6.1){$X_5$}
\put(6.4,5.4){\vector(-1, -1){1.3}}    
\put(6.6,5.4){\vector(0,-1){3.1}}  
\put(6.4,5.6){\vector(-1,0){1.3}} 
\put(6.4,5.8){\line(-1,1){1.3}}    
\put(1.2,5.6){\circle*{0.2}}
\put(1.2,6.1){$X_3$}
\put(1.2,3.8){\circle*{0.2}}
\put(1.2,4.3){$X_1$}
\put(1.4,3.8){\line(1,0){1.3}}
\put(3,3.8){\circle*{0.2}} 
\put(2.4,4.3){$X_2$}
\put(3.2,3.8){\vector(1,0){1.3}} 
\put(4.7,3.8){\circle*{0.2}} 
\put(4.4,4.3){$X_6$}
\put(4.9,3.6){\vector(1,-1){1.3}}    
\end{picture}
\end{center}
\caption{A partially oriented graph due to Principle V 
(arrow  $(X_{10},X_9)$). Nodes $X_1$,  $X_3$, $X_8$, $X_9$ are legitimately 
removable.} \label{abbsechs}
\end{figure}
} 
\newcommand{\AbbSieben}
{
\begin{figure}
\begin{center}
\begin{picture}(8,8.3)
\put(4.7,2){\circle*{0.2}} 
\put(4.4,1.3){$X_7$}
\put(4.7,2){\vector(0, 1){1.3}}    
\put(4.9,2.2){\line(1, 2){1.7}}    
\put(3,5.6){\circle*{0.2}}
\put(2.9,6.1){$X_4$} 
\put(4.7,7.4){\circle*{0.2}} 
\put(5,8.1){$X_{10}$} 
\put(4.5,7.2){\line(-1,-1){1.3}}    
\put(6.6,5.6){\circle*{0.2}}
\put(6.6,6.1){$X_5$}
\put(6.4,5.4){\vector(-1, -1){1.3}}    
\put(6.4,5.8){\line(-1,1){1.3}}    
\put(1.2,3.8){\circle*{0.2}}
\put(1.2,4.3){$X_1$}
\put(1.4,3.8){\line(1,0){1.3}}
\put(3,3.8){\circle*{0.2}} 
\put(2.4,4.3){$X_2$}
\put(3.2,3.8){\vector(1,0){1.3}} 
\put(4.7,3.8){\circle*{0.2}} 
\put(4.4,4.3){$X_6$}
\end{picture}
\end{center}
\caption{After legitimate removal of nodes $X_3$, $X_8$ and $X_9$.
(The arrow $(X_3,X_4)$ was inforced).
 Nodes $X_1$,  $X_4$, $X_6$ are legitimately 
removable.} \label{abbsieben}
\end{figure}
} 
\newcommand{\AbbAcht}
{
\begin{figure}
\begin{center}
\begin{picture}(8,8.3)
\put(1.0,2.5){\circle*{0.2}} 
\put(1.2,2.7){$X_{11}$}
\put(1.2,2.5){\vector(1,0){1.5}} 
\put(1.2,2.7){\vector(1,1){1.5}} 
\put(1.2,2.3){\vector(1,-1){1.5}} 
\put(1.0,4.5){\circle*{0.2}} 
\put(1.2,4.7){$X_{16}$}
\put(3.0,0.5){\circle*{0.2}} 
\put(3.2,0.3){$X_{13}$}
\put(3.2,0.5){\vector(1,0){1.5}} 

\put(3.0,2.5){\circle*{0.2}} 
\put(3.2,2.7){$X_{14}$}
\put(3.0,2.7){\line(0,1){1.5}} 
\put(3.0,2.3){\line(0,-1){1.5}} 
\put(3.0,4.5){\circle*{0.2}} 
\put(3.2,4.7){$X_{15}$}
\put(2.8,4.5){\vector(-1,0){1.5}} 
\put(5.0,2.5){\circle*{0.2}} 
\put(5.2,2.7){$X_{12}$}
\put(4.8,2.5){\vector(-1,0){1.5}} 
\put(4.8,2.7){\vector(-1,1){1.5}} 
\put(4.8,2.3){\vector(-1,-1){1.5}} 
\put(5.0,0.5){\circle*{0.2}} 
\put(5.2,0.7){$X_{17}$}
\end{picture}
\end{center}
\caption{A partially oriented graph. Active trails and active p-trails.
Consider the p-trail $(X_{11},X_{15})$, $(X_{15},X_{12})$ and assume the node $X_{13}$ being blocked. The p-trail under consideration is active. 
In any derived dag either the trail
 $(X_{11},X_{15})$, $(X_{15},X_{12})$ 
 is active  or there exists another trail
(e.g.  $(X_{11},X_{13})$, $(X_{13},X_{12})$) that is active. 
} \label{abbacht}
\end{figure}
} 
\newcommand{\AbbNeun}
{
\begin{figure}
\begin{center}
\begin{picture}(8,3.1)
\put(1.0,0.5){\circle*{0.2}} 
\put(1.2,0.5){\vector( 1,0){1.5}} 
\put(1.2,0.7){$X_i$}
\put(3.0,0.5){\circle*{0.2}} 
\put(3.2,0.7){$X_j$}
\put(5.0,0.5){\circle*{0.2}} 
\put(5.2,0.7){$X_k$}
\put(4.8,0.5){\vector(-1,0){1.5}} 
\put(3.0,2.5){\circle*{0.2}} 
\put(3.2,2.7){$X_l$}
\put(2.8,2.3){\line(-1,-1){1.5}} 
\put(3.0,2.3){\line( 0,-1){1.5}} 
\put(3.2,2.3){\line( 1,-1){1.5}} 
\end{picture}
\end{center}
\caption{Visualisation to the Proof of the Theorem on Principle V         
} \label{abbneun}
\end{figure}
} 
\begin{document}

\machetitel


\newcommand{\V}{{\bf V }}
\newcommand{\SS}{{\bf S }}



\begin{abstract}
Spirtes, Glymour and Scheines \cite{Spirtes:90b} formulated a Conjecture
that a direct dependence test and a head-to-head meeting test
would suffice to construe directed acyclic graph decompositions of 
a joint probability distribution (Bayesian network)
 for which
Pearl's d-separation \cite{Geiger:90} applies. 
This Conjecture was later shown to be a direct consequence of 
a result of Pearl and Verma \cite{Verma:Pearl:90}, cited as Theorem 1 in 
\cite{Pearl:Verma:91}, see also Theorem 3.4. in \cite{Spirtes:93}).

This paper is intended to prove this Conjecture in a new way, by 
exploiting   the concept of p-d-separation (partial dependency separation) \cite{Klvopotek:2000} . 
While Pearl's d-separation works with 
Bayesian networks, p-d-separation is 
intended to apply to causal networks: that is partially oriented networks in 
which orientations are given to only to those edges, that  express 
statistically confirmed causal influence, whereas undirected edges express 
existence of direct influence without possibility of determination of 
direction of causation.

As a consequence of the particular way of proving the validity of this 
Conjecture, 
an algorithm  for construction of all the 
directed acyclic graphs (dags) carrying the available independence 
information is also presented. 
The notion of a partially oriented graph (pog) is introduced 
and 
within 
this graph the notion of p-d-separation is defined. 
It is demonstrated that the  p-d-separation within the pog is equivalent 
to d-separation in all derived dags. 
%
\end{abstract}

\section{Introduction}

An analysis detecting only a model of joint probability distribution of a set 
of variables is not of itself a reliable guide to judgements about policy, 
which inevitably involves causal conclusions. The policy implications of 
empirical data can be completely reversed by alternative hypotheses about the 
causal relationships of variables and the estimates of a particular causal 
influence can be radically  altered by changes in the assumptions made about 
other dependencies. For these reasons one of the common aims of empirical 
research in the social sciences is to determine the causal relations among a 
set of variables, and to estimate the relative importance 
of various causal factors \cite{Spirtes:90b}. 

How can causal relations among variables be discovered ?\\
The difficulty of this question may be realized if one considers the number 
of possible causal models for a given set of variables. If the causal 
dependence of one variable on the other is represented by a directed edge in a 
graph, then there are 
$4^{n\cdot (n-1)/2}$ such models for a set of n variables. If causal cycles are 
forbidden, then the number of possible (acyclic) graph models is still 
immense: (always more than $2^{n\cdot (n-1)/2}$) for 12 variables it 
is:521~939~651~343~829~405~020~504~063 \cite{Hilary:73}. Even if the ordering 
of the variables consistent with the given
directed acyclic graph (dag) is known, the number of 
possibilities remains large: for 12 variables it is $7\cdot 10^{19}$.  

A scientist addressing problem areas where causal questions are of concern is 
therefore faced with an extremely difficult discovery problem, for which three 
avenues of solution  can be mentioned: (i) use  experimental controls to 
eliminate  most of alternative causal structures (ii) introduce prior 
knowledge to restrict the space of alternatives; and (iii) use features of 
sample data to restrict the space of alternatives. Following the first avenue 
may be too  expensive or even not feasible, the second one may be not 
advisable if theoretical foundations are too vague so that restrictions imposed 
by them may prevent from discovering  the true underlying causal structure.

As far as 
the third alternative is concerned, methodologists routinely warn against 
such  inferences (exploiting the slogan "correlation does not imply 
causation"), warn that "substantial knowledge", not sample data, should 
determine the causal structure of a model 
(compare e.g. \cite{Leohlin:87,James:82}).
Procedures that use the sample data 
are denounced as "data mining" or "ransacking".

Bayesian networks 
(called also belief networks, probabilistic networks)
encode properties of probability distributions using 
directed acyclic graphs (dag). Their usage is spread among many disciplines 
such as Artificial Intelligence \cite{Pearl:88}, Decision Analysis 
\cite{Howard:81,Shachter:88}, Economics \cite{Wold:64}, Genetics 
\cite{Wright:34}, Philosophy \cite{Glymour:87}, and Statistics 
\cite{Laurizen:88,Smith:87}.

Bayesian networks are popular due to existence of numerous
efficient 
 methods of 
reasoning with probabilities if the joint probability distribution
has an underlying dag structure
\cite{Pearl:86,Pearl:88,Shachter:90,%
Shachter:90b,Shenoy:90}.

Spirtes, Glymour and Scheines \cite{Spirtes:90b} formulated a Conjecture
(called below SGS Conjecture)
that a direct dependence test and a head-to-head meeting test
would suffice to construe directed acyclic graph decompositions
from data  of 
a joint probability distribution (Bayesian  network)
 for which
Pearl's d-separation \cite{Geiger:90} applies.
\NS{This conjecture was later shown to be a direct consequence of 
a result of Pearl and Verma \cite{Verma:Pearl:90}, cited as Theorem 1 in 
\cite{Pearl:Verma:91}, see also Theorem 3.4. in \cite{Spirtes:93}).\\
}

 This paper is intended
to prove the SGS Conjecture in a new way, by 
exploiting the concept of p-d-separation (partial dependency separation). 
While Pearl's d-separation, indirectly referred to in the SGS Conjecture, 
 works with 
Bayesian networks, p-d-separation is 
intended to apply to causal networks: that is partially oriented networks in 
which orientations are given  only to those edges, that  express 
statistically 
suggested causal influence, whereas undirected edges express 
existence of direct influence without possibility of determination of 
direction of causation. 

The concept of p-d-separation seems to be more natural in the context of the 
SGS Conjecture,
because the direct dependence test and the head-to-head meeting test cannot in  
general recover all edge orientations in the dag to be reconstructed from the data.  
Hence at a point the construction of a dag requires 
an arbitrary (though compatible) orientation of the unoriented edges.  
It is only after this arbitrary edge orientation step that Pearl's d-separation  
concept can be applied to reason qualitatively about conditional dependence and  
independence of variables.   
However, usage of arbitrary edge orientation in order to reason about independence  
seems to be very strange. We suspected that the partially oriented graph (pog) just  
before arbitrary orientation may as well be used 
for qualitative reasoning about (in)dependence  without inserting  
unsupported information about edge orientation. Hence the concept of 
p-d-separation has been explained..

We have still to keep in mind that not any partially oriented graph is suitable for  
reasoning about independence. Too few edge orientation information may produce 
 to misleading results. One has to incorporate also the
information about the misses that is that some head-to-head orientations have been  
rejected.  By formulating so-called principles II$^c$, IV and V 
we managed to pass enough edge orientation information into partially oriented graph  
(pog), 
obtained using  Spirtes et al.  direct 
dependence test and  head-to-head meeting test to orient the pog, 
so that p-d-separation in the resulting pog is equivalent 
to Pearl's d-separation in any arbitrarily derived compatible dag.


As a consequence of validity of the SGS Conjecture and the particular proof based on  
p-d-separation, 
an algorithm for construction of all the 
directed acyclic graphs (dags) carrying the available independence 
information is also presented and justified. 

\section{A review of d-separation and its important properties}

Let us first recall the definition of a Bayesian network, its relation to 
intrinsic 
causal networks and the important d-separation properties shared by both.

\begin{df}  \cite{Geiger:90}
 A 
{\em Bayesian network } is a pair (D,P) where D is a dag
(directed acyclic graph)
 and P is a 
probability 
distribution called the {\em underlying distribution}. Each node i in D 
corresponds to a variable $X_i$  in P, a set of nodes I corresponds to a set 
of variables $X_I$ and $x_i, x_I$
 denote values drawn from the domain of $X_i$ 
 and from the (cross product) domain of $X_I$ respectively. Each node in the 
network is  regarded as a storage cell for the distribution 
$P(x_i | x_{\pi (i)})$ where $X_{\pi (i)}$ is a set of nodes corresponding to
the 
parent nodes $\pi(i)$ of i.  The underlying distribution represented by a 
Bayesian network is computed via:
$$P(x_1,...,x_n) = \prod_{i=1}^{n} P(x_i  |  x_{\pi (i)})$$
\end{df}

Formally, the Bayesian network is nothing more than a representation of a
joint probability distribution. In practice, however, the edges within the dag  
associated with the Bayesian network and their orientation are 
intuitively understood as expression of direct causal links and the expression of  
orientation of causation. As it is in general assumed that a variable can cause  
itself  neither directly nor indirectly, we can impose the partial ordering of  
variables such that for a variable $X$ the set of its direct predecessors $\pi(X)$  
completely determines the value of $X$ up to a noise, that disturbs the value of  
$X$ ($X=f(\pi(X))+\epsilon$). Under these circumstances obviously the joint  
probability distribution in all the 
variables can be appropriately expressed by a Bayesian network with its dag  
representing exactly this partial ordering of causation. 

Of course, a Bayesian network can be transformed into another more complex Bayesian  
network (e.g. by edge reversal \cite{Shachter:90b}) that expresses the same joint  
probability distribution. However, the resulting more complex Bayesian network will  
no more be the
intrinsic causal network. Hence the intrinsic causal network is intuitively  
associated with the simplest Bayesian network. 
A more formal definition of causation and causal networks will be given in the next  
section.

The Bayesian networks and hence also the intrinsic        causal networks
have several important properties. One of them is the so-called
d-separation, relating statistical (conditional) independence to some 
graph-theoretic properties of the dag associated with the 
Bayesian network. 
We cite in this section subsequently large portions of section 1 and 2 of 
Geiger et al. \cite{Geiger:90}.\\

\begin{df}
A {\em trail } in a dag is a sequence of links that form a path in the 
 underlying undirected graph. A node $\beta$ is called a {\em head-to-head 
node}  with 
respect to a trail t if there are two consecutive links $\alpha \rightarrow 
\beta$ and $\beta \leftarrow \gamma$ on that t. 
\end{df}

\begin{df}
A trail t connecting nodes $\alpha$ and $\beta$ is said to be {\em active } 
given a set of nodes L, if (1) every head-to-head-node wrt t either is or has 
a descendent in L and (2) every other node on t is outside L. Otherwise t is 
said to be {\em blocked } (given L).
\end{df}

\begin{df}
If J,K and L are three disjoint sets of nodes in a dag D, then L is said to 
{\em d-separate } J from K, denoted $I(J,K | L)_D$  iff no trail between a
node 
in J and a node in K is active given L.
\end{df}

It has been shown in \cite{Geiger:90b} that 
\begin{thm}
Let L be a set of nodes in a dag D, and let $\alpha,\beta \notin L$ be two 
additional nodes in D. Then $\alpha$ and $\beta$ are connected via an active 
trail  (given L) iff  $\alpha$ and $\beta$ are connected via a simple (i.e. 
not possessing cycles in the underlying undirected graph) active trail (given 
L).
\end{thm}

\begin{df}
If $X_J,X_K,X_L$ are three disjoint sets of variables of a distribution P, 
then $X_J,X_K$ are said to be conditionally independent given $X_L$ (denoted 
$I(X_J,X_K  | X_L)_P$ iff $P(x_J,x_K | x_L)=P(x_J | x_L)
\cdot  P(x_K | x_L)$ for
all 
possible values of $X_J,X_K,X_L$ for which $P(x_L)>0$. 
$I(X_J,X_K  | X_L)_P$ is called a {\em 
(conditional) independence statement}
\end{df}

\begin{thm}
Let $P_D=\{P | $(D,P) is a Bayesian network\}. Then:\\

$I(J,K | L)_D$ iff $I(X_J,X_K  | X_L)_P$ for all $P \in P_D$.
\end{thm}

The "only if" part (soundness) states that whenever  $I(J,K | L)_D$ holds in
D, 
it must represent an independence that holds in every underlying distribution. 

The "if" part (completeness) asserts that any independence that is not 
detected by d-separation cannot be shared by all distributions in $P_D$ and 
hence cannot be revealed by non-numeric methods. 


\section{The SGS Conjecture}

Many writers have connected causation with statistical dependence. In 
\cite{Spirtes:90b} the following 
understanding of causation 
was assumed: 
\begin{df}
"Let \V be a set of random variables with a joint probability distribution. 
We say that variables X,Y $\in$\V are
 {\em directly causally dependent} if and 
 only if there is a causal dependency between X,Y (either the value of X 
influences the value of Y or the value of Y influences the value of X or the 
value of a third variable not in \V influences the values of both X and Y)
that does not involve any other variable in \V.
\end{df}

\begin{df}
"We  say that {\em B is directly causally dependent on A} provided that A and 
B 
are causally dependent and the direction of causal influence is from A to 
B."
\end{df}

As self-causation of variables is discarded by Spirtes, Glymour and Scheines, we  
assume that: 
\begin{df}
An {\em intrinsic causal network } is a directed acyclic graph (dag) over the set of  
variables \V such that  there is an edge 
connecting  variables X,Y $\in$\V if and only if 
 X,Y $\in$\V are
  directly causally dependent and the edge is oriented from X to Y 
if and only if  Y is directly causally dependent on X. 
We additionally assume that the joint probability P in variables \V 
is accessible for computation of marginals and conditional marginals.
\end{df}

 In 
\cite{Spirtes:90b} the following principles for association of causation with  
statistical 
dependence  were assumed:

"{\bf Principle I: }  For all X,Y in \V, X and Y are directly causally 
dependent 
if and only if for every subset \SS of \V not containing X or Y, X and Y are 
not statistically independent conditional on \SS" (page 185)\\

"{\bf Principle II: } if A and B are directly causally dependent and B and C 
are directly causally dependent, but A and C are not, then:
B is causally dependent on A, and B is causally dependent on C if and only if 
A and C are statistically dependent conditional on any set of variables 
containing B and not containing A or C." (pages 186-187).\\

"{\bf Principle III: } A directed acyclic graph represents a probability 
distribution on the variables that are vertices of the graph if and only if\\
for all vertices X,Y and all sets \SS   of vertices in the graph
(X,Y $\notin$ \SS), \SS 
d-separates 
X and Y if and only if X and Y are independent conditional on \SS" (page 193). 
\\

We refrain here from citing the d-separation definition presented therein,
as it is semantically a bit different from that of Geiger and Pearl 
\cite{Geiger:90}. and we are convinced that the latter is the correct one, so 
we  cited the latter in the previous section. \\

Spirtes et al. claim the following:
\begin{thm} \label{iiipi} 
"Let P be a probability distribution represented by an acyclic directed graph 
G according to Principle III. Then G is an orientation (G has the undirected 
structure) of an undirected graph U that represents P according to Principle 
I."
\end{thm}
\begin{thm}  \label{iiipii} 
"Principle III implies Principle II."
\end{thm}

{\bf SGS Conjecture: } "Let $\Gamma$ be the set of directed 
graphs 
that represent probability distribution P according to Principle III. Then 
$\Gamma$ is also the set of directed graphs obtained from P by Principles I 
and II."\\

From the above-mentioned theorems we can easily guess that 
the orientation of edges in the intrinsic 
causal network may not be accessible to our observation even if the 
causation mechanism fits the statistical assumptions. Therefore, for 
practical reasons, we need the concept of a causal network that takes this 
into account.  
\begin{df}
A {\em  causal network } is a 
graph
 over the set of variables \V as the set of node labels 
 with some edges unoriented and other oriented,
that can be transformed to a dag by proper orienting the unoriented edges, 
such that \\
(1)  there is an edge 
connecting  variables X,Y $\in$\V if and only if 
 X,Y $\in$\V are
  directly causally dependent and\\
(2)
if the edge connecting X and Y is oriented then 
 the edge is oriented from X to Y 
if and only if  Y is directly causally dependent on X. \\
We additionally assume that the joint probability P in variables \V 
is accessible for computation of marginals and conditional marginals.
\end{df}

The intrinsic causal network is a special case of  causal network  in which all the  
edges are oriented.

\section{From the SGS conjecture to a theorem}

We shall stress at this point the 
immense  importance of the Spirtes et 
al. Conjecture. 
Principle III, when applied for construction of the dag, refers for every
pair of nodes to the whole future dag structure. Hence it may prove quite 
cumbersome to apply - virtually nearly  any possible dag needs to be 
checked.

On the other hand, Principle I refers in the construction stage only to the 
pair of nodes (and to other nodes), but never checks any future (directed or 
undirected)  edges of 
the dag. Principle II refers only to nodes in the "neighbourhood" due to the 
Principle I and it refers only to nodes and to two already established 
undirected edges and never to the future directed structure of the dag.

Let us introduce some notions. First let us define a {\em partially oriented 
graph} (pog). 
\begin{df}
 A {\em partially oriented graph} (pog)
is a structure (\V,E,O),
where 
\begin{enumerate}
\item 
  \V is  the set of nodes, 
\item 
E is    the set of 
edges with an edge being a subset of \V with cardinality 2, 
\item 
O:E$\rightarrow 2 ^ {V \times V}$ is    the 
orientation function of edges assigning each edge $\{X_i,X_j\}$ in E
\begin{enumerate}
\item 
 either 
the orientation $\{\}$ (no orientation)
\item 
 or  $\{(X_i,X_j)\}$  (from $X_i$ to 
$X_j$),
\item 
 or   $\{(X_j,X_i)\}$  (from $X_j$ to 
$X_i$) 
\item 
or   $\{(X_i,X_j),(X_j,X_i)\}$  (both from $X_i$ to 
$X_j$ ) and from $X_j$ to  $X_i$). 
\end{enumerate}
\end{enumerate}
\end{df}

 The  last (bidirectional)  orientation  is  an 
unpleasant one, 
but may occur in processes described below, If the first (empty) orientation 
is assigned, the edge is called unoriented, otherwise it is called   
oriented.

Furthermore let us call two edges {\em neighbouring edges} iff they share a 
vertex.   Let $\{X_i,X_j\}$ and  $\{X_k,X_j\}$ be neighbouring edges (they 
share $X_j$ so they are neighbouring at $X_j$). We call them {\em bridged 
edges} iff  there exists an edge  $\{X_i,X_k\}$ in E. Otherwise they are 
called {\em unbridged}. The edge  $\{X_i,X_j\}$  (with respect to the 
neighbouring pair of edges) is said to be {\em head-to-neighbour}
 oriented iff  $(X_i,X_j) \in O( \{X_i,X_j\})$.   The edge    
$\{X_i,X_j\}$ 
(with respect to the 
neighbouring pair of edges) is said to be {\em tail-to-neighbour}
 oriented iff  $(X_j,X_i) \in O( \{X_i,X_j\})$.  

As the
 first step in proving the validity of the SGS conjecture let us strengthen 
 Theorem \ref{iiipi}. 
In general,   several different dags $G_1, G_2, ... $
may  represent the probability distribution P according to Principle III. 
Each of the dags has an underlying structure (unoriented graph) $U_1,U_2,...$. 
But for a given set of variables, 
it is easily seen that  Principle I applied to a 
distribution P yields exactly one undirected graph U, 
because in the formulation of Principle I there is no reference to the
structure of the underlying dag. 
Hence U$=U_1=U_2=...$. 
So the phrase  "an undirected graph U" should be replaced with "the  
undirected graph U"  in  Theorem \ref{iiipi}.  

\AbbEins

(So if the intrinsic graph is given by Fig.\ref{abbeins} then Principle I 
yields a graph given by Fig.\ref{abbzwei}).\\

\AbbZwei

Let us look at this theorem more closely. If two nodes/variables $X_i$ and 
$X_j$ are connected via an undirected edge within the U-graph generated by 
Principle I, then there exists no set of variables $Y_1,....,Y_k$ such that 
for every combination of values $P(x_i,x_j  |  y_1,...,y_n)= P(x_i  |  
y_1,...,y_n) \cdot P(x_j  |  y_1,...,y_n)$ as otherwise the edge would not be 
inserted. Assume for a moment Principle III would not generate a directed 
edge connecting both variables in a directed graph D. Then in this graph D 
a d-separation of both variables can be found: take simply the set of nodes 
that  directly precede any of the variables. But this would enforce 
conditional independence in contradiction with the result established 
previously. So any edge generated by Principle I is also present in every 
graph generated by Principle III. \\
On the other hand if Principle I establishes that there is no undirected edge 
connecting both variables then there exists a set of variables on which these 
two are conditionally independent. But then Principle III cannot establish an 
edge between them as there would exist no d-separation between them. So 
whenever Principle I establishes no edge between variables, no edge will be 
established by Principle III.

Let us consider the graph U generated by Principle I. Let us consider 
partial orientations of the  graph U generated from it by Principle II. It is 
easily seen that there may be only one such orientation. Let us turn our 
attention to  Theorem  \ref{iiipii}. 
Let us consider a head-to-head meeting of directed edges $(X_i,X_l)$, 
$(X_j,X_l)$ generated by Principle II, that is $X_i,X_j$ not being 
directly 
connected in U,  $X_i,X_l$ being directly connected in U, 
$X_j,X_l$  being directly connected in U, no set containing $X_l$ rendering 
$X_i,X_j$ independent. Then Principle III has also to generate this 
head-to-head meeting as the existence of the trail of directed edges  
$(X_i,X_l)$, $(X_j,X_l)$ guarantees in this case that no d-separation 
 containing $X_l$ exists.  So every head-to-head-meeting generated by 
Principle 
II occurs also in every graph generated by Principle III. On the other hand, 
if during testing  independence by means of Principle II for the edges  
$(X_i,X_l)$, $(X_j,X_l)$  a set containing $X_l$ was detected such that it  
renders $X_i,X_j$ independent, then head-to-head meeting of these edges 
must not occur if Principle III is applied.

In this way we have established that: if there exists a dag of the 
distribution generated by Principle III, then application of Principles I 
and II will deliver its undirected structure and orientation of all  those 
unbridged pairs of 
arcs that  meet head-to-head at a node. 
(So if the intrinsic graph is given by Fig.\ref{abbeins} then Principle II 
yields a graph given by Fig.\ref{abbdrei}).

\AbbDrei

Let us now discuss which orientations of other arcs are established rigidly 
by Principle III. Pearl's definition of d-separation refers to arc 
orientation at following nodes:
(1) head-to-head nodes 
(2) direct and indirect descendants of head-to-head nodes 

So let us establish the following principle:\\

{\bf Principle II$^c$} Let H be a partially oriented graph 
generated by Principles I and II. 
Whenever  
  $\{X_i,X_j\}$ and  $\{X_k,X_j\}$ are neighbouring unbridged edges, with
 $\{X_i,X_j\}$ being head-to-neighbour oriented and    $\{X_k,X_j\}$ being 
unoriented, orient    $\{X_k,X_j\}$  tail-to-neighbour. \\

Please notice that Principle  II$^c$ is a kind of operationalization of 
Principle II, as it is a direct consequence of the "if and  only if" 
expression in   Principle II. It has been introduced because the 
formulation of Principle II directs our attention to orienting edges 
head-to-head, but it is less obvious that it also implies some head-to-tail 
orientations.

Obviously, the following theorem holds: 
\begin{thm}  \label{iiipiic} 
Principle III implies Principle II$^c$.
\end{thm}

The Theorem is obvious if we consider the previous ones.
(So if the intrinsic graph is given by Fig.\ref{abbeins} then Principle  
II$^c$ yields a graph given by Fig.\ref{abbvier}).

\AbbVier

Furthermore let us propose the following principle:

{\bf Principle IV: } Let  H be a partially oriented graph.
 Let the subgraph H' 
of H contain only oriented edges in H. Let $\{X_i,X_j\}$  be an unoriented 
edge in H. If $X_j$ is a descendent of $X_i$ in H', then orient this edge 
from $X_i$ to $X_j$. \\

Obviously
\begin{thm}  \label{iiipiv} 
Dag-structure and Principle III imply Principle IV.
\end{thm}

(So if the intrinsic graph is given by Fig.\ref{abbeins} then Principle IV 
yields a graph given by Fig.\ref{abbfuenf}).\\

\AbbFuenf

{\bf Principle V:} Let  H be a partially oriented graph 
generated by Principles I and II.  
Let the unbridged edges $\{X_i,X_j\}$, $\{X_k,X_j\}$
be oriented head-to-head by  Principle II. Let 
both edges $\{X_i,X_l\}$, $\{X_j,X_l\}$ or both edges 
 $\{X_k,X_l\}$, 
 $\{X_j,X_l\}$, or all the edges 
 $\{X_i,X_l\}$,
 $\{X_k,X_l\}$, 
 $\{X_j,X_l\}$
 be left unoriented in the 
process. Then orient  $\{X_j,X_l\}$ as  from $X_l$ to $X_j$.\\

(If the intrinsic graph is given by Fig.\ref{abbeins} then Principle V  
yields a graph given by Fig.\ref{abbsechs}).\\

\AbbSechs

\begin{thm}  \label{iiipv} 
Dag-structure and Principle III imply Principle V.
\end{thm}

\AbbNeun

\AnfBeweis
The edges 
 $\{X_i,X_l\}$, $\{X_k,X_l\}$
(see Fig.\ref{abbneun})
 are unbridged (because  $\{X_i,X_j\}$, 
$\{X_k,X_j\}$ are unbridged), hence their orientation head-to-head is 
 excluded (as Principle II didn't orient them). Hence either we have 
orientation $( X_l,X_i )$ or  $( X_l,X_k )$\\
 Let us assume the orientation
 $ (X_l,X_i)$ of  $\{X_i,X_l\}$. Then if  $\{X_l,X_j\}$ would be oriented  
$( X_j,X_l )$ then $X_j,X_l,X_i$ would form an oriented  cycle, hence H would 
not be a dag. So this is impossible.\\
 Let us assume the orientation
 $ (X_l,X_k)$ of  $\{X_k,X_l\}$. Then if  $\{X_l,X_j\}$ would be oriented  
$( X_j,X_l )$ then $X_j,X_l,X_k$ would form an oriented  cycle, hence H would 
not be a dag. So this is impossible.\\
Hence $\{X_l,X_j\}$ must  be oriented  
$( X_l,X_j )$ \\
\EndBeweis

To prove 
the SGS conjecture 
 we shall introduce first the notion of p-d-separation.

\begin{df}
A {\em p-trail } in a pog is a sequence of links that form a path in the 
underlying  undirected graph. A node $\beta$ is called a head-to-head node 
with 
respect to a p-trail t if there are two consecutive links $\alpha \rightarrow 
\beta$ and $\beta \leftarrow \gamma$ on that t. 
A p-trail is minimal iff no two of its  succeeding links on the p-trail are 
bridged in the graph. 
\end{df}

\begin{df}
A p-descendent of a node n in a pog is any node m such that there exists a 
minimal p-trail from n to m such that every oriented link on the p-trail is 
oriented from n to m and an oriented edge (m,n) does not exist in the graph.
 \end{df}

\begin{df}
A p-trail t connecting nodes $\alpha$ and $\beta$ is said to be {\em active } 
given a set of nodes L, if (1) every head-to-head-node wrt t either is or has 
 a p-descendent in L and (2) every other node on t is outside L. Otherwise t 
is 
said to be {\em blocked } (given L).
\end{df}

\begin{df}
If J,K and L are three disjoint sets of nodes in a pog H, then L is said to 
{\em p-d-separate } J from K, denoted $I(J,K | L)_H$  iff no minimal
p-trail between a 
node in J and a node in K is active given L.
\end{df}

We  claim that 
\begin{thm}
Let L be a set of nodes in a pog H, and let $\alpha,\beta \notin L$ be two 
additional nodes in H. Then $\alpha$ and $\beta$ are connected via an active 
p-trail  (given L) iff  $\alpha$ and $\beta$ are connected via a simple (i.e. 
not possessing cycles in the underlying undirected graph) active p-trail 
(given L).
\end{thm}


Now let us formulate the central theorem of this paper. 

\begin{thm} \label{iuiipiii}
Let D be a dag generated by Principle III. 
Let H be a pog generated by Principles I, II, II$^c$, IV and V.
Then $I(J,K | L)_H$ iff $I(J,K  | L)_D$ 
\end{thm}

\AbbAcht

\AnfBeweis
To show this, let us consider an  active minimal p-trail. We claim that there 
exists then an active trail. \\
If after final orientation no head-to-head meeting occurs on the p-trail then 
this is also the interesting active trail. Otherwise if there exists a 
head-to-head-meeting on the underlying trail then two cases are possible:
(1) it existed on the original p-trail, (2) it did not exist on the original 
p-trail. The second case is impossible since then it must have been generated 
 by Principle  II (the meeting edges are unbridged). So we have had also a 
head-to-head-meeting on the original p-trail. So let us consider the 
p-descenders of the head-to-head-meeting. No head-to-head-meeting could have 
been generated on the path as the p-trail to the descendent  was minimal. 
p-descendents
of head-to-head meetings
 connected by unoriented links  form a kind of equivalence class in that if 
the edges (A,B), (C,B)  are there and D is a p-descendent of B on a totally 
unoriented path then oriented edges (A,D) and (C,D) are also present. So 
p-descendants are either descendants (OK) or are   such              
     predecessors, that they form together with the nodes of the primary 
p-trail but the discussed head-to-head node
       a minimal p-trail containing that predecessor as a head-to-head node 
and which proves to be an active trail in the dag (see Fig.\ref{abbacht}).\\
Let us consider an active minimal trail. We claim that then there exists a 
minimal p-trail. First of all, all the successors are also p-successors. 
Second, a minimal trail is also a minimal p-trail. Now the question  is 
whether 
or not it is also active. As the trail is minimal, no head-to-head meeting 
will vanish on the p-trail. Hence also the successor requirement is met. So 
the proof is complete.
\EndBeweis

This theorem actually corresponds straight forwardly to the 
SGS conjecture. The only difference to it is the extensive use of 
Principle II$^c$ that  is actually a kind of exploitation of the Spirtes 
Principle II. Furthermore, it is to
 some extent constructive: it states how it 
 is possible to uncover the d-separations applying only Principles I, II,  
II$^c$, IV 
and V for construction of a pog, and without actually instantiating a single 
dag. It is immediately visible, that any dag compatible with the pog expresses 
exactly all the independences  Principle III dag does and hence is a 
Principle III dag.

We can however be still more constructive and formulate the construction
algorithm for generation of all the dags according to Principle III
based only on the results of Principles  I, II,  
II$^c$, IV 
and V and the definition of a dag.

Let us define the legitimate removal of a node from the pog graph:
\begin{df}
{\em A node can be removed legitimately from a pog } iff all the oriented 
edges it 
meets are oriented towards it, and all pairs of 
edges meeting at it for which at 
least one is unoriented, are bridged. \\
\end{df}

{\large \bf Pog-to-dag algorithm:}\\
\begin{itemize}
\item[1.] Find  a legitimately removable node in the pog, remove it with edges 
meeting it while marking the edges as oriented towards this node. 
\item[2.] Proceed with Step.1 until all the nodes are removed.
\item[3.] Orient the edges of the original pog so as they were marked in step
1.
\end{itemize}

Notice that the algorithm is non-deterministic: At step 1 we can have several  
candidate legitimately removable nodes.  Selecting any of them may lead to different,  
though statistically equivalent dags.

\AbbSieben

(Compare Fig.\ref{abbsechs}, Fig.\ref{abbsieben}).
We claim that:

\begin{thm} 
Let there exist a dag obtainable from  Principle III.
Let G be a pog generated from Principles I, II, II$^c$, IV 
and V.  Then every dag 
obtained from the pog B by the above algorithm 
is a Principle III dag. Every Principle III dag for this population is a dag 
obtainable from G by means of the above algorithm.  
\end{thm}
\AnfBeweis
This is easily seen as on the  one hand every dag has a legitimately 
removable node, and on  the other hand the orientations generated by the 
above algorithm do  not  lead  to  any  conflict  with  Principles 
I,II,II$^c$, IV and V, 
if a dag exists.  \\
\EndBeweis

In this way we hope to have also shown the   validity of the SGS
conjecture 
definitely, giving a constructive algorithm to generate the dag out of a pog 
which is necessary for belief network applications. \\


\NS{The result of this section may be stated as follows}

\begin{thm} 
Let $\Gamma$ be the set of directed 
graphs 
that represent probability distribution P according to Principle III. Then 
$\Gamma$ is also the set of directed graphs obtained from P by Principles I 
and II.
 \end{thm}%
\AnfBeweis
Let us look closely at Theorem \ref{iuiipiii}.
From Theorems \ref{iiipi} and \ref{iiipii} we know that any 
dag D in $\Gamma$ must have been generated also by Principles I and II.
 As Principles II$^c$, 
IV and V follow from Principles I and II and from the property of being a dag  
(look at Theorems \ref{iiipiic}, \ref{iiipiv},    \ref{iiipv}), then any dag 
in  $\Gamma$ as generated by Principle III would also be 
generated by Principles I, II, II$^c$, IV and V. Let us take now any of these 
dags in $\Gamma$, say D. Let us assume that from the respective pog H 
generated 
by Principles I, II, II$^c$, IV and V 
(that is in fact from the only such pog H)
a different dag D' may be derived 
beside 
D. From Theorem  \ref{iuiipiii} we have: $I(J,K | L)_H$ iff $I(J,K  | L)_D$,
but 
also:  $I(J,K | L)_H$ iff $I(J,K  | L)_{D'}$.  Hence also  $I(J,K | L)_D$ iff 
$I(J,K  | L)_{D'}$. But then D' must also have been generated by Principle III
as 
both D and D' carry the same independence information.\\
 So we see immediately that any dag in $\Gamma$ must have been generated by 
Principles I and II and all the dags derived via Principles I and II must be 
in $\Gamma$.
\EndBeweis

\section{Discussion and Conclusions}

\ML{
In this paper, the SGS conjecture has been proven to be a theorem.
This implies that it is  possible to infer causal structure from 
data  if this structure has the form of a directed acyclic graph. Strictly 
speaking: 
The statistical inference allows for deducing a set of such candidate causal 
structures with indication which fragment of the causal structure is shared by 
all the candidates. 
}

\NS{ In this paper, the notion of p-d-separation was explained for causal 
networks, paralleling the notion of d-separation of \cite{Geiger:90} in belief 
networks. Its usefulness and power for representation of 
dependence/independence relations in causal networks was demonstrated by 
providing another proof of the SGS conjecture from 
\cite{Spirtes:90b}. }

Specifically: In this paper, new Principles II$^c$, IV and V were introduced 
allowing to orient constructively
more edges of the undirected underlying graph of the causal 
structure than it was possible using only original Principles I and II of 
Spirtes et al  \cite{Spirtes:90b}. Furthermore, an algorithm was given 
allowing 
for derivation of all the dags having identical dependence/independence 
information as the partially oriented graph derived from Principles I and II, 
provided at least one dag exists.
 The notion of p-d-separation paralleling d-separation 
of Geiger, Verma  and Pearl \cite{Geiger:90}, being applicable to partially 
oriented graphs was presented and has been shown to carry the same 
dependence/independence information as all the d-separations  of all 
compatible dags.

Over the last years a number of alternative (both general and specialized) 
methods for construction of probabilistic belief networks has been proposed  
(compare the method described in  \cite{Cooper:92} and other 
discussed in last 
sections therein
and also in \cite{Spirtes:93}). 
The SGS conjecture investigated here deserved 
special 
attention because it relates the oriented structure of a directed acyclic 
graph representation to the causal relationship in the described part of 
reality. 
\ML{
 Two essential complementary conclusions can be drawn from proving 
SGS  Conjecture in this paper: (i) if one recovers a dag structure for the 
probability distribution one derives more than just a formal description and 
(ii) for proper construction of a dag causation is essential.

Further research on the subject is needed, especially concerning 
approximations binding combinatorial explosion with the number of variables 
considered. }

With the power of p-d-separation, a (partially recovered) causal network 
structure can be used for qualitative reasoning about statistical 
dependence/independence in just the same manner as a belief network structure 
is exploited for qualitative statistical reasoning in \cite{Geiger:90}. 

We shall draw attention to the fact that 
the concept of p-d-separation 
appears  to be more natural in the context of the 
SGS conjecture,
because Principles I and II do not in general recover all edge orientations in the  
dag to be reconstructed from the data. Therefore, to obtain a dag 
an arbitrary (though compatible) orientation of the unoriented edges
is needed, because   
 only after this arbitrary edge orientation step  Pearl's 
d-separation concept can be applied to reason qualitatively about marginal 
and conditional dependence and independence of variables.   
But such an orientation stage adds information not supported by anything and hence  
it is unnatural. 
p-d-separation allows for the same qualitative reasoning  but without
the arbitrary information. 
Also the edge-orientation stage may turn out to be complex because 
not all edge orientations may turn out to be consistent both with dag
properties and constraints resulting from principle II. So back-tracking
while randomly orienting may be necessary. 
Checking p-d-separation in a partially oriented graph may be less
cumbersome. 

 Not any partially oriented graph is suitable for reasoning about independence. Too  
few edge orientation information may lead to misleading results. 
  By formulating so-called principles II$^c$, IV and V 
we managed to pass enough edge orientation information into partially oriented graph  
(pog), 
obtained using  Spirtes et al.  Principles I and II, 
so that p-d-separation in the resulting pog is equivalent 
to Pearl's d-separation in any arbitrarily derived compatible dag.

This research is restricted to the case of causally sufficient sets of variables  
that is to cases when all variables needed to construct a dag of the intrinsic  
underlying process are available.  
Further research is needed to extend the notion of p-d-separation on
causal network recovery from data under causal insufficiency, that is
whenever influential variables remain  hidden.
The concept of Possible-D-Sep from \cite{Spirtes:93} can be considered as a good  
starting point, but the need to maintain information
(on discarded head-to-head meetings)
 beside the  partially oriented graph is to some extent discouraging. 
It should be attempted to get the outside constraints into the partially oriented  
graphs as done in this paper for causally sufficient cases.


As the only reference to the data in this methodology relies on conditional 
dependence/independence test, also an investigation has been started on 
possibilities of 
extension of the methodology onto Dempsterian-Shaferian belief networks and 
other constructs for which the dependence/independence test from the data may 
be carried out.

\end{document}